\documentstyle[aclap]{article}

\input{epsf}

\title{\vspace{-0.5in}Memory-Based Learning: Using Similarity for
Smoothing}  

\author{Jakub Zavrel \and Walter Daelemans\\
Computational Linguistics\\
Tilburg University\\
PO Box 90153\\
5000 LE Tilburg\\
The Netherlands\\
{\tt \{zavrel,walter\}@kub.nl}}

\begin{document}
\maketitle
\bibliographystyle{acl}

\vspace{-0.5in}

\begin{abstract} 
This paper analyses the relation between the use of similarity in
Memory-Based Learning and the notion of backed-off smoothing in
statistical language modeling. We show that the two approaches are
closely related, and we argue that feature weighting methods in the
Memory-Based paradigm can offer the advantage of automatically
specifying a suitable domain-specific hierarchy between most specific
and most general conditioning information without the need for a large
number of parameters. We report two applications of this approach:
PP-attachment and POS-tagging. Our method achieves state-of-the-art
performance in both domains, and allows the easy integration of
diverse information sources, such as rich lexical representations.
\end{abstract}

\section{Introduction}

Statistical approaches to disambiguation offer the advantage of making
the most likely decision on the basis of available evidence. For this
purpose a large number of probabilities has to be estimated from a
training corpus.  However, many possible conditioning events are not
present in the training data, yielding zero Maximum Likelihood (ML)
estimates. This motivates the need for {\em smoothing} methods, which
re-estimate the probabilities of low-count events from more reliable
estimates.

Inductive generalization from observed to new data lies at the heart
of machine-learning approaches to disambiguation. In Memory-Based
Learning\footnote{The Approach is also referred to as Case-based,
Instance-based or Exemplar-based.} (MBL) induction is based on the use
of similarity \cite{StanfillWaltz,aha-ibl,Cardie94,Daelemans95}. In
this paper we describe how the use of similarity between patterns
embodies a solution to the sparse data problem, how it relates to
backed-off smoothing methods and what advantages it offers when
combining diverse and rich information sources.

We illustrate the analysis by applying MBL to two tasks where
combination of information sources promises to bring improved
performance: PP-attachment disambiguation and Part of Speech tagging.

\section{Memory-Based Language Processing}

The basic idea in Memory-Based language processing is that processing
and learning are fundamentally interwoven. Each language experience
leaves a memory trace which can be used to guide later
processing. When a new instance of a task is processed, a set of
relevant instances is selected from memory, and the output is produced
by analogy to that set.

The techniques that are used are variants and extensions of the
classic $k$-nearest neighbor ($k$-NN) classifier algorithm. The
instances of a task are stored in a table as patterns of feature-value
pairs, together with the associated ``correct'' output. When a new
pattern is processed, the $k$ nearest neighbors of the pattern are
retrieved from memory using some similarity metric. The output is then
determined by extrapolation from the $k$ nearest neighbors, i.e. the
output is chosen that has the highest relative frequency among the
nearest neighbors.

Note that no abstractions, such as grammatical rules, stochastic
automata, or decision trees are extracted from the examples. Rule-like
behavior results from the linguistic regularities that are present in
the patterns of usage in memory in combination with the use of an
appropriate similarity metric.  It is our experience that even limited
forms of abstraction can harm performance on linguistic tasks, which
often contain many subregularities and exceptions~\cite{Daelemans96}.

\subsection{Similarity metrics}

The most basic metric for patterns with symbolic features is the {\bf
Overlap metric} given in equations~\ref{distance} and~\ref{overlap};
where $\Delta(X,Y)$ is the distance between patterns $X$ and $Y$,
represented by $n$ features, $w_{i}$ is a weight for feature $i$, and
$\delta$ is the distance per feature. The $k$-NN algorithm with this
metric, and equal weighting for all features is called {\sc ib1}
\cite{aha-ibl}. Usually $k$ is set to 1.

\begin{equation}
\Delta(X,Y) = \sum_{i=1}^{n}\ w_{i}\ \delta(x_{i},y_{i})
\label{distance}
\end{equation}

where:\\
\begin{equation}
\delta(x_{i}, y_{i}) = 0\ if\ x_{i} = y_{i},\ else\ 1
\label{overlap}
\end{equation}

This metric simply counts the number of (mis)matching feature values
in both patterns. If we do not have information about the importance of
features, this is a reasonable choice. But if we do have some
information about feature relevance one possibility would be to add
linguistic bias to weight or select different
features~\cite{Cardie96}. An alternative---more empiricist---approach,
is to look at the behavior of features in the set of examples used for
training. We can compute statistics about the relevance of features by
looking at which features are good predictors of the class
labels. Information Theory gives us a useful tool for measuring
feature relevance in this way~\cite{Quinlan86,Quinlan93}.\\

{\bf Information Gain} (IG) weighting looks at each feature in
isolation, and measures how much information it contributes to our
knowledge of the correct class label. The Information Gain of feature
$f$ is measured by computing the difference in uncertainty
(i.e. entropy) between the situations without and with knowledge of
the value of that feature (Equation~\ref{IGgain}).

\begin{equation}
w_{f} = H(C) - \frac{ \sum_{v \in V_{f}} P(v) \times H(C|v)}{si(f)}
\label{IGgain}
\end{equation}

\begin{equation}
si(f) = - \sum_{v \in V_{f}} P(v) \log_{2} P(v)
\label{splitinfo}
\end{equation}

Where $C$ is the set of class labels, $V_{f}$ is the set of values for
feature $f$, and $H(C) = - \sum_{c \in C} P(c) \log_{2} P(c)$ is
the entropy of the class labels. The probabilities are estimated from
relative frequencies in the training set. The normalizing factor
$si(f)$ (split info) is included to avoid a bias in favor of features
with more values. It represents the amount of information needed to
represent all values of the feature (Equation~\ref{splitinfo}). The
resulting IG values can then be used as weights in
equation~\ref{distance}. The $k$-NN algorithm with this metric is
called {\sc ib1-ig}~\cite{daelemans-hyphen}.\\

The possibility of automatically determining the relevance of features
implies that many different and possibly irrelevant features can be
added to the feature set. This is a very convenient methodology if
theory does not constrain the choice enough beforehand, or if we wish
to measure the importance of various information sources
experimentally.

Finally, it should be mentioned that MB-classifiers, despite their
description as table-lookup algorithms here, can be implemented to
work fast, using e.g.~tree-based indexing into the
case-base~\cite{Daelemans+al97}. 

\section{Smoothing of Estimates}

The commonly used method for probabilistic classification (the
Bayesian classifier) chooses a class for a pattern $X$ by picking the
class that has the maximum conditional probability $P(class|X)$. This
probability is estimated from the data set by looking at the relative
joint frequency of occurrence of the classes and pattern $X$. If
pattern $X$ is described by a number of feature-values $x_{1}, \dots
,x_{n}$, we can write the conditional probability as $P(class|x_{1},
\dots ,x_{n})$. If a particular pattern $x'_{1}, \dots ,x'_{n}$ is not
literally present among the examples, all classes have zero ML
probability estimates. Smoothing methods are needed to avoid zeroes on
events that could occur in the test material.

There are two main approaches to smoothing: count re-estimation
smoothing such as the Add-One or Good-Turing
methods~\cite{ChurchGale}, and Back-off type methods
\cite{Bahl+al,Katz,ChenGoodman,Samuelsson}. We will focus here on a
comparison with Back-off type methods, because an experimental
comparison in~\newcite{ChenGoodman} shows the superiority of Back-off
based methods over count re-estimation smoothing methods.  With the
Back-off method the probabilities of complex conditioning events are
approximated by (a linear interpolation of) the probabilities of more
general events:

\begin{eqnarray}
\tilde{p}(class|X) & = & \lambda_{X} \hat{p}(class|X)  + \lambda_{X'} \hat{p}(class|X') \nonumber \\
		   &   & + \cdots + \lambda_{X^{n}} \hat{p}(class|X^{n})
\label{linear_interpolation}
\end{eqnarray}

Where $\tilde{p}$ stands for the smoothed estimate, $\hat{p}$ for the
relative frequency estimate, $\lambda$ are interpolation weights,
$\sum_{i=0}^{n} \lambda_{X^{i}} = 1$, and $X \prec X^{i}$ for all $i$,
where $\prec$ is a (partial) ordering from most specific to most
general feature-sets\footnote{$X \prec X'$ can be read as $X$ is more
specific than $X'$.}
(e.g the probabilities of trigrams ($X$) can be approximated by
bigrams ($X'$) and unigrams ($X''$)).  The weights of the linear
interpolation are estimated by maximizing the probability of held-out
data (deleted interpolation) with the forward-backward algorithm. An
alternative method to determine the interpolation weights without
iterative training on held-out data is given in Samuelsson (1996).

We can assume for simplicity's sake that the $\lambda_{X^{i}}$ do not
depend on the value of $X^{i}$, but only on $i$. In this case, if $F$
is the number of features, there are $2^{F}-1$ more general terms, and
we need to estimate $\lambda_{i}$'s for all of these.  In most
applications the interpolation method is used for tasks with clear
orderings of feature-sets (e.g. n-gram language modeling) so that many
of the $2^{F}-1$ terms can be omitted beforehand. More recently, the
integration of information sources, and the modeling of more complex
language processing tasks in the statistical framework has increased
the interest in smoothing
methods~\cite{CollinsBrooks,Ratnaparkhi,Magerman,NgLee,Collins}.  For
such applications with a diverse set of features it is not
necessarily the case that terms can be excluded beforehand.

If we let the $\lambda_{X^{i}}$ depend on the value of $X^{i}$, the
number of parameters explodes even faster. A practical solution for
this is to make a smaller number of {\em buckets} for the $X^{i}$,
e.g. by clustering (see e.g. \newcite{Magerman}). 

Note that linear interpolation (equation~\ref{linear_interpolation})
actually performs two functions. In the first place, if the most
specific terms have non-zero frequency, it still interpolates them
with the more general terms. Because the more general terms should
never overrule the more specific ones, the $\lambda_{X^{i}}$ for the
more general terms should be quite small. Therefore the interpolation
effect is usually small or negligible. The second function is the pure
back-off function: if the more specific terms have zero frequency, the
probabilities of the more general terms are used instead. Only if
terms are of a similar specificity, the $\lambda$'s truly serve to
weight relevance of the interpolation terms.

If we isolate the pure back-off function of the interpolation equation
we get an algorithm similar to the one used in~\newcite{CollinsBrooks}.
It is given in a schematic form in Table~\ref{back-off-CB}. Each step
consists of a back-off to a lower level of specificity. There are as
many steps as features, and there are a total of $2^F$ terms, divided
over all the steps. Because all features are considered of equal
importance, we call this the {\em Naive Back-off} algorithm.

\begin{table}[htb]
If $f(x_{1},...,x_{n})>0$:\\
\vspace{5mm}
$\tilde{p}(c|x_{1},...,x_{n}) = 
\frac{f(c,x_{1},...,x_{n})}{f(x_{1},...,x_{n})}$\\
Else if $f(x_{1},...,x_{n-1},\ast)+...+f(\ast,x_{2},...,x_{n})>0$:\\
\vspace{5mm}
$\tilde{p}(c|x_{1},...,x_{n}) = 
\frac{f(c,x_{1},...,x_{n-1},\ast)+... +f(c,\ast,x_{2},...,x_{n})}
     {f(x_{1},...,x_{n-1},\ast)+...+f(\ast,x_{2},...,x_{n})}$\\
Else if \dots:\\
\vspace{5mm}
$\tilde{p}(c|x_{1},...,x_{n}) = \frac{\cdots}{\cdots}$\\
Else if $f(x_{1},\ast,...,\ast)+...+f(\ast,...,\ast,x_{n})>0$:\\
\vspace{5mm}
$\tilde{p}(c|x_{1},...,x_{n}) = 
\frac{f(c,x_{1},\ast,...,\ast)+...+f(c,\ast,...,\ast,x_{n})}
     {f(x_{1},\ast,...,\ast)+ ... + f(\ast,...,\ast,x_{n})}$\\
\caption{The Naive Back-off smoothing algorithm. $f(X)$ stands for the
frequency of pattern $X$ in the training set. An asterix ($\ast$)
stands for a wildcard in a pattern. The terms at a higher level in the
back-off sequence are more specific ($\prec$) than the lower levels.}
\label{back-off-CB}
\end{table}

Usually, not all features $x$ are equally important, so that not all
back-off terms are equally relevant for the re-estimation. Hence, the
problem of fitting the $\lambda_{X^{i}}$ parameters is replaced by a
term selection task.  To optimize the term selection, an evaluation of
the up to $2^F$ terms on held-out data is still necessary.  In
summary, the Back-off method does not provide a principled and
practical domain-independent method to adapt to the structure of a
particular domain by determining a suitable ordering $\prec$ between
events. In the next section, we will argue that a formal
operationalization of similarity between events, as provided by MBL,
can be used for this purpose.  In MBL the similarity metric and
feature weighting scheme automatically determine the implicit back-off
ordering using a domain independent heuristic, with only a few
parameters, in which there is no need for held-out data.

\section{A Comparison}

\begin{figure*}[htb]
        \begin{center}
		\leavevmode
                \epsfxsize=10cm
                \epsfysize=8cm
                \epsffile{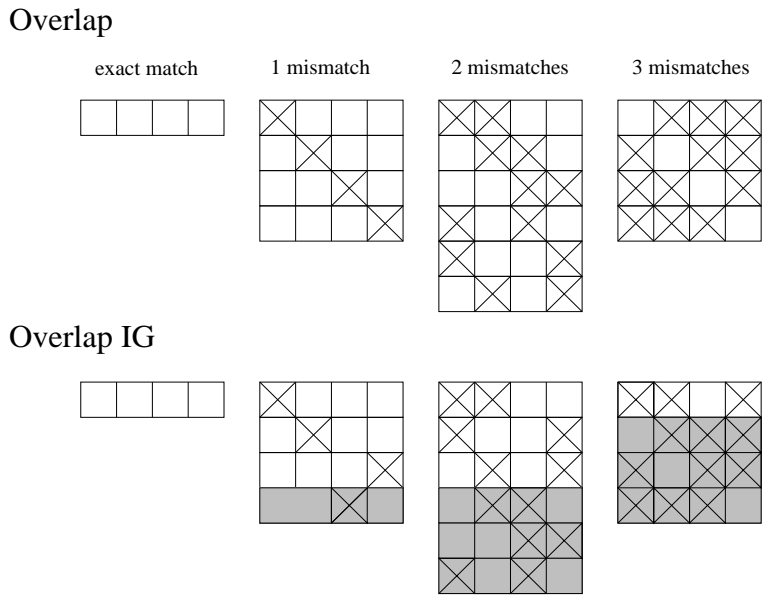}
		\caption{An analysis of nearest neighbor sets into
		buckets (from left to right) and schemata (stacked). IG
		weights reorder the schemata. The grey schemata are
		not used if the third feature has a very high weight
		(see section 5.1).}
		\label{buckets}
	\end{center}
\end{figure*}

If we classify pattern $X$ by looking at its nearest neighbors, we are
in fact estimating the probability $P(class|X)$, by looking at the
relative frequency of the class in the set defined by $sim_{k}(X)$,
where $sim_{k}(X)$ is a function from $X$ to the set of most similar
patterns present in the training data\footnote{Note that MBL is not
limited to choosing the best class. It can also return the conditional
distribution of all the classes.}. Although the name ``$k$-nearest
neighbor'' might mislead us by suggesting that classification is based
on exactly $k$ training patterns, the $sim_{k}(X)$ function given by
the Overlap metric groups varying numbers of patterns into {\em
buckets} of equal similarity. A bucket is defined by a particular
number of mismatches with respect to pattern $X$. Each bucket can
further be decomposed into a number of {\em schemata} characterized by
the position of a wildcard (i.e. a mismatch).  Thus $sim_{k}(X)$
specifies a $\prec$ ordering in a Collins \& Brooks style back-off
sequence, where each bucket is a step in the sequence, and each schema
is a term in the estimation formula at that step. In fact, the
unweighted overlap metric specifies exactly the same ordering as the
Naive Back-off algorithm (table~\ref{back-off-CB}). In
Figure~\ref{buckets} this is shown for a four-featured pattern. The
most specific schema is the schema with zero mismatches, which
corresponds to the retrieval of an identical pattern from memory, the
most general schema (not shown in the Figure) has a mismatch on every
feature, which corresponds to the entire memory being best neighbor.

If Information Gain weights are used in combination with the Overlap
metric, individual schemata instead of buckets become the steps of the
back-off sequence\footnote{Unless two schemata are exactly tied in
their IG values.}.  The $\prec$ ordering becomes slightly more
complicated now, as it depends on the number of wildcards {\em and} on the
magnitude of the weights attached to those wildcards.  Let $S$ be the
most specific (zero mismatches) schema. We can then define the $\prec$
ordering between schemata in the following equation, where
$\Delta(X,Y)$ is the distance as defined in equation~\ref{distance}.

\begin{equation}
S' \prec S'' \ \Leftrightarrow \ \Delta(S,S') < \Delta(S,S'')
\label{subset_with_IG}
\end{equation}

Note that this approach represents a type of implicit parallelism. The
importance of the $2^{F}$ back-off terms is specified using only $F$
parameters---the IG weights--, where $F$ is the number of
features. This advantage is not restricted to the use of IG weights;
many other weighting schemes exist in the machine learning literature
(see ~\newcite{Wettschereck+al97} for an overview).\\

Using the IG weights causes the algorithm to rely on the most specific
schema only. Although in most applications this leads to a higher
accuracy, because it rejects schemata which do not match the most
important features, sometimes this constraint needs to be weakened.
This is desirable when: (i) there are a number of schemata which are
almost equally relevant, (ii) the top ranked schema selects too few
cases to make a reliable estimate, or (iii) the chance that the few
items instantiating the schema are mislabeled in the training material
is high. In such cases we wish to include some of the lower-ranked
schemata. For case (i) this can be done by discretizing the IG weights
into bins, so that minor differences will lose their significance, in
effect merging some schemata back into buckets. For (ii) and (iii),
and for continuous metrics~\cite{StanfillWaltz,CostSalzberg} which
extrapolate from exactly $k$ neighbors\footnote{Note that the schema
analysis does not apply to these metrics.}, it might be necessary to
choose a $k$ parameter larger than 1. This introduces one additional
parameter, which has to be tuned on held-out data. We can then use the
distance between a pattern and a schema to weight its vote in the
nearest neighbor extrapolation. This results in a back-off sequence in
which the terms at each step in the sequence are weighted with respect
to each other, but without the introduction of any additional
weighting parameters. A weighted voting function that was found to
work well is due to~\newcite{Dudani}: the nearest neighbor schema
receives a weight of 1.0, the furthest schema a weight of 0.0, and
the other neighbors are scaled linearly to the line between these two
points.

\section{Applications}

\subsection{PP-attachment}

In this section we describe experiments with MBL on a data-set of
Prepositional Phrase (PP) attachment disambiguation cases. The problem
in this data-set is to disambiguate whether a PP attaches to the verb
(as in {\em I ate pizza with a fork}) or to the noun (as in {\em I ate
pizza with cheese}). This is a difficult and important problem,
because the semantic knowledge needed to solve the problem is very
difficult to model, and the ambiguity can lead to a very large number
of interpretations for sentences.

We used a data-set extracted from the Penn Treebank WSJ corpus
by~\newcite{Ratnaparkhi+al}. It consists of sentences containing the
possibly ambiguous sequence {\em verb noun-phrase PP}. Cases were
constructed from these sentences by recording the features: verb, head
noun of the first noun phrase, preposition, and head noun of the noun
phrase contained in the PP. The cases were labeled with the attachment
decision as made by the parse annotator of the corpus. So, for the two
example sentences given above we would get the feature vectors {\tt
ate,pizza,with,fork,V.} and {\tt ate,pizza,with,cheese,N.} The
data-set contains 20801 training cases and 3097 separate test cases,
and was also used in~\newcite{CollinsBrooks}.\\

The IG weights for the four features ({\tt V,N,P,N}) were respectively
0.03, 0.03, 0.10, 0.03. This identifies the preposition as the most
important feature: its weight is higher than the sum of the other
three weights. The composition of the back-off sequence following from
this can be seen in the lower part of Figure~\ref{buckets}. The
grey-colored schemata were effectively left out, because they include
a mismatch on the preposition.

\begin{table}
\begin{center}
\begin{tabular}{|l|l|}
\hline
Method          & \% Accuracy\\
\hline
{\sc ib1} (=Naive Back-off)	& 83.7 \%\\
{\sc ib1-ig} 	& 84.1 \%\\
LexSpace IG	 & 84.4 \%\\
\hline
Back-off model (Collins \& Brooks) & 84.1 \%\\
C4.5 (Ratnaparkhi et al.)          & 79.9 \%\\
Max Entropy (Ratnaparkhi et al.)& 81.6 \%\\
Brill's rules (Collins \& Brooks)  & 81.9 \%\\
\hline
\end{tabular}
\end{center}
\caption{Accuracy on the PP-attachment test set.}
\label{PPattach}
\end{table}

Table~\ref{PPattach} shows a comparison of accuracy on the test-set of
3097 cases. We can see that {\sc ib1}, which implicitly uses the same
specificity ordering as the Naive Back-off algorithm already performs
quite well in relation to other methods used in the
literature.
Collins \& Brooks' (1995) Back-off model is more sophisticated than
the naive model, because they performed a number of validation
experiments on held-out data to determine which terms to include and,
more importantly, which to exclude from the back-off sequence. They
excluded all terms which did not match in the preposition! Not
surprisingly, the 84.1\% accuracy they achieve is matched by the
performance of {\sc ib1-ig}. The two methods exactly mimic each others
behavior, in spite of their huge difference in design. It should
however be noted that the computation of IG-weights is many orders of
magnitude faster than the laborious evaluation of terms on held-out
data.

We also experimented with rich lexical representations obtained in an
unsupervised way from word co-occurrences in raw WSJ text
~\cite{ZavrelVeenstra95,Schuetze}. We call these representations
Lexical Space vectors. Each word has a numeric 25 dimensional vector
representation. Using these vectors, in combination with the IG
weights mentioned above and a cosine metric, we got even slightly
better results. Because the cosine metric fails to group the patterns
into discrete schemata, it is necessary to use a larger number of
neighbors ($k=50$). The result in Table~\ref{PPattach} is obtained
using Dudani's weighted voting method.

Note that to devise a back-off scheme on the basis of these
high-dimensional representations (each pattern has $4\times25$
features) one would need to consider up to $2^{100}$ smoothing terms.
The MBL framework is a convenient way to further experiment with
even more complex conditioning events, e.g. with semantic labels
added as features.

\subsection{POS-tagging}

Another NLP problem where combination of different sources of
statistical information is an important issue, is POS-tagging,
especially for the guessing of the POS-tag of words not present in the
lexicon. Relevant information for guessing the tag of an unknown word
includes contextual information (the words and tags in the context of
the word), and word form information (prefixes and suffixes, first and
last letters of the word as an approximation of affix information,
presence or absence of capitalization, numbers, special characters
etc.). There is a large number of potentially informative features
that could play a role in correctly predicting the tag of an unknown
word ~\cite{Ratnaparkhi,Weischedel+al,Daelemans+al96}.  A
priori, it is not clear what the relative importance is of these
features.

We compared Naive Back-off estimation and MBL with
two sets of features:
\begin{itemize}
\item
{\sc pdass}: the first letter of the unknown word (p), the tag of the
word to the left of the unknown word (d), a tag representing the set
of possible lexical categories of the word to the right of the unknown
word (a), and the two last letters (s). The first letter provides
information about capitalisation and the prefix, the two last letters
about suffixes.
\item
{\sc pdddaaasss}: more left and right context features, and more
suffix information.
\end{itemize}

\begin{figure}[h]
        \begin{center}
		\leavevmode
                \epsfxsize=8cm
                \epsfysize=6cm
                \epsffile{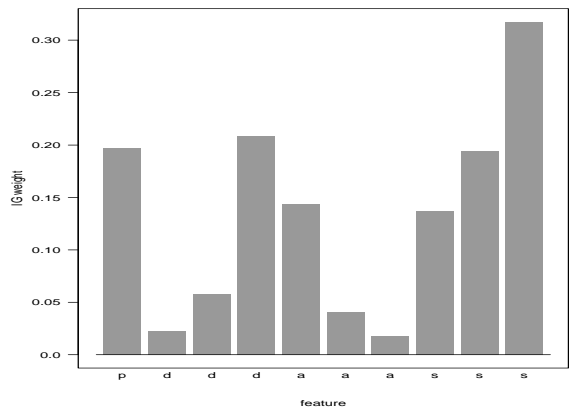}
		\caption{IG values for features used in predicting the
		tag of unknown words.}
		\label{tagig}
	\end{center}
\end{figure}

The data set consisted of 100,000 feature value patterns taken from
the Wall Street Journal corpus. Only open-class words were used during
construction of the training set. For both {\sc ib1-ig} and Naive Back-off,
a 10-fold cross-validation experiment was run using both {\sc pdass}
and {\sc pdddaaasss} patterns. The results are in Table
\ref{tagres}. The IG values for the features are given in Figure
\ref{tagig}.

\begin{table}[b]
\begin{center}
\begin{tabular}{|l|l|l|}
\hline
 & {\sc ib1}, Naive Back-off & {\sc ib1-ig} \\
\hline
{\sc pdass} & 88.5 (0.4) & 88.3 (0.4)\\
{\sc pdddaaasss} & 85.9 (0.4) & 89.8 (0.4)\\
\hline
\end{tabular}
\end{center}
\caption{Comparison of generalization accuracy of Back-off and
Memory-Based Learning on prediction of category of unknown words. All
differences are statistically significant (two-tailed paired t-test,
$p<0.05$). Standard deviations on the 10 experiments are between
brackets.}\label{tagres}
\end{table}

The results show that for Naive Back-off (and {\sc ib1}) the addition
of more, possibly irrelevant, features quickly becomes detrimental
(decrease from 88.5 to 85.9), even if these added features do make a
generalisation performance increase possible (witness the increase
with {\sc ib1-ig} from 88.3 to 89.8). Notice that we did not actually
compute the $2^{10}$ terms of Naive Back-off in the {\sc pdddaaasss}
condition, as IB1 is guaranteed to provide statistically the same
results.  Contrary to Naive Back-off and {\sc ib1}, memory-based
learning with feature weighting ({\sc ib1-ig}) manages to integrate
diverse information sources by differentially assigning relevance to
the different features. Since noisy features will receive low IG
weights, this also implies that it is much more noise-tolerant.

\section{Conclusion}

We have analysed the relationship between Back-off smoothing and
Memory-Based Learning and established a close correspondence between
these two frameworks which were hitherto mostly seen as unrelated. An
exception is the use of similarity for alleviating the sparse data
problem in language
modeling~\cite{EssenSteinbiss92,Brown+al92,Dagan+al94}. However, these
works differ in their focus from our analysis in that the emphasis is
put on similarity between {\em values} of a feature (e.g. words),
instead of similarity between patterns that are a (possibly
complex) combination of many features. 

The comparison of MBL and Back-off shows that the two approaches
perform smoothing in a very similar way, i.e. by using estimates from
more general patterns if specific patterns are absent in the training
data. The analysis shows that MBL and Back-off use exactly the same
type of data and counts, and this implies that MBL can safely be
incorporated into a system that is explicitly probabilistic. Since the
underlying $k$-NN classifier is a method that does not necessitate any
of the common independence or distribution assumptions, this promises
to be a fruitful approach.
 
A serious advantage of the described approach, is that in MBL the
back-off sequence is specified by the used similarity metric, without
manual intervention or the estimation of smoothing parameters on
held-out data, and requires only one parameter for each feature
instead of an exponential number of parameters.  With a
feature-weighting metric such as Information Gain, MBL is particularly at an
advantage for NLP tasks where conditioning events are complex, where
they consist of the fusion of different information sources, or when
the data is noisy. This was illustrated by the experiments on
PP-attachment and POS-tagging data-sets.

\section*{Acknowledgements}

This research was done in the context of the ``Induction of Linguistic
Knowledge'' research programme, partially supported by the Foundation
for Language Speech and Logic (TSL), which is funded by the
Netherlands Organization for Scientific Research (NWO). We would like
to thank Peter Berck and Anders Green for their help with software for
the experiments.

\end{document}